\documentclass[lettersize,journal]{IEEEtran}
\usepackage{amsmath,amsfonts}
\usepackage{algorithmic}
\usepackage{algorithm}
\usepackage{array}
\usepackage[caption=false,font=normalsize,labelfont=sf,textfont=sf]{subfig}
\usepackage{textcomp}
\usepackage{stfloats}
\usepackage{url}
\usepackage{verbatim}
\usepackage{graphicx}
\usepackage{cite}
\usepackage{orcidlink}
\usepackage{booktabs}

\providecommand{\new}[1]{#1}

\begin{document}

\title{Urban Risk-Aware Navigation via VQA-Based Event Maps for People with Low Vision}

\author{Antoni~Valls~\orcidlink{0009-0004-5511-2795}, Jorge~Tyrakowski~\orcidlink{0009-0002-5561-343X}, My-Dinh Le, Alex Cabañeros, Wen-Huang Cheng~\orcidlink{0000-0002-4662-7875}, Jordi~Sanchez-Riera~\orcidlink{0000-0002-4803-5742}%

\thanks{This work has been submitted to the IEEE for possible publication. Copyright may be transferred without notice, after which this version may no longer be accessible.}%
\thanks{Antoni Valls is with the Universitat Polit\`ecnica de Catalunya and with the Institut de Rob\`otica i Inform\`atica Industrial (IRI), CSIC-UPC (e-mail: antoni.valls.cifre@upc.edu).}
\thanks{Jordi Sanchez-Riera is with the Institut de Rob\`otica i Inform\`atica Industrial (IRI), CSIC-UPC (e-mail: jsanchez@iri.upc.edu).}%
\thanks{Jorge Tyrakowski, My-Dinh Le and Wen-Huang Cheng are with the Department of Computer Science and Information Engineering, National Taiwan University, Wen-Huang Cheng is also with VinUniversity (e-mail: jorgetyrakowski@cmlab.csie.ntu.edu.tw, lenguyenhoangmydinh@gmail.com, wenhuang@csie.ntu.edu.tw).}%
\thanks{Alex Cabañeros is with Biel Glasses S.L. (e-mail: alex.cabaneros@bielglasses.com).}}

\markboth{}
{IEEE TRANSACTIONS ON INTELLIGENT TRANSPORTATION SYSTEMS}%


\maketitle

\begin{abstract}

\new{Urban navigation can be challenging for people with low vision, as pedestrians, potholes, trees and other sidewalk elements may pose difficult-to-detect hazards. In this work, we propose a VQA-based framework that leverages Vision-Language Models (VLMs) to generate risk event maps for offline pedestrian route risk assessment, rather than real-time walking instruction generation. A three-level hierarchical query structure enables fine-grained scene understanding without task-specific retraining, with each question generating a structured event describing an environmental element or potential hazard and its associated risk score. These events are spatially aggregated through a weighted scoring mechanism to classify street segments into four risk bands, producing risk event maps that highlight potentially hazardous route segments. We introduce a geographically diverse dataset spanning 89 cities across all continents, comprising over 4,800 annotated images and 95,000 answered questions. We benchmark five VQA models---ViLT, LLaVA, InstructBLIP, Qwen-VL, and Gemma-4---for urban hazard detection and risk assessment, with Gemma-4 achieving the best overall performance (F1-score of 0.71). We also conduct a user study with 20 participants using urban scenes captured with smart glasses across Barcelona, evaluating the agreement between participants' risk perceptions and the model's predictions, as well as their trust in its assessments.}

\end{abstract}


\section{Introduction}\label{sec:intro}

\begin{figure}[t!] 
    \centering
    \includegraphics[width=1\linewidth]{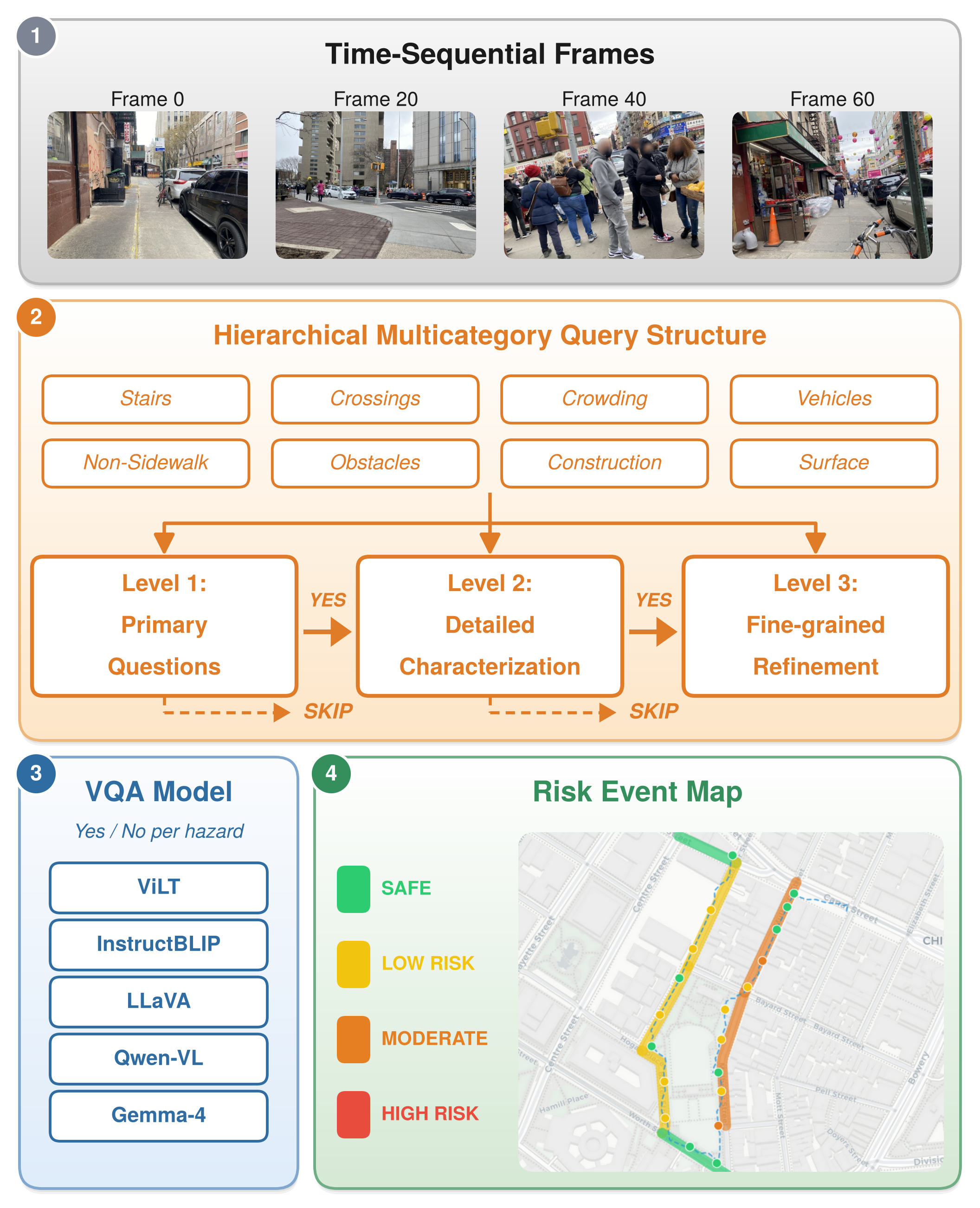}
    \caption{Overview of the proposed framework. Time-sequential images extracted from video streams are fed into a three-level Hierarchical Multicategory Query Structure (orange), where Level-1 primary questions trigger conditional follow-up queries at Levels 2 and 3, progressively refining hazard characterization across eight pedestrian safety categories. Queries are processed by \new{five VQA model architectures---ViLT, LLaVA, InstructBLIP, Qwen-VL, and Gemma-4}---which produce binary Yes/No responses per hazard category (blue). Model outputs are aggregated via a weighted risk scoring formulation and spatially mapped into a risk event map (green), color-coded into four risk bands: \new{\textit{Safe}, \textit{Low Risk}, \textit{Moderate}, and \textit{High Risk}.}}
    \label{fig:pipeline}
\end{figure}

\IEEEPARstart{V}{isual} impairment represents a major global challenge affecting millions of individuals worldwide. As of 2015, approximately 253 million people were living with visual impairment, including 36 million who were blind and 217 million with moderate to severe vision loss \cite{ackland2018world}. Unlike common refractive errors such as myopia or astigmatism that can be readily corrected with glasses, these numbers reflect functional vision loss that significantly impacts daily activities and quality of life. Projections suggest that the number of blind and visually impaired individuals could reach 703 million by 2050 , driven by population aging and growth \cite{ackland2018world}. This demographic shift creates an urgent need for innovative assistive technologies that can support independent navigation and enhance the safety of \new{people with low vision} in their daily activities.

\new{For people with low vision, navigating urban environments presents substantial challenges that directly impact mobility and independence. Obstacles along sidewalks, unmarked stairs, approaching vehicles, crowded areas, and construction zones can be difficult to anticipate, compromising pedestrian safety. Wearable devices such as smartphones and smart glasses offer promising platforms for assistive navigation, traditionally focusing on real-time hazard detection and guidance. However, assessing pedestrian routes in advance offers complementary advantages, allowing potential hazards to be identified before a journey begins, risk information to be shared among users, and alternative routes between destinations to be compared based on their safety. Such an approach could support more informed route selection, helping people with low vision anticipate challenging segments and identify safer routes to their intended destinations.}

\new{Early assistive navigation systems commonly relied on task-specific vision algorithms to detect obstacles and environmental features, such as stairs or pedestrian crossings \cite{teng,wangDeepLeaningbasedUltrafast2022,hwang2024safecrossinterpretablerisk}. Although effective for targeted tasks, these approaches are constrained by predefined detection capabilities, making it difficult to flexibly characterize diverse or unforeseen urban hazards. Recent advances in Vision-Language Models (VLMs) offer a more general alternative by enabling visual scenes to be interpreted through natural-language queries, reducing reliance on dedicated detection pipelines for individual tasks.}

\new{VLMs and Visual Question Answering (VQA) combine visual perception and natural-language reasoning to interpret scenes and answer context-dependent questions \cite{PARK2023100548,vqa}. Recent works have explored their application to assistive navigation providing visual reasoning for real-time walking assistance through contextual instructions \cite{yuanWalkVLMAidVisually2025} and reminders \cite{Werby-RSS-24,tasko}. While these approaches demonstrate the potential of VLMs for flexible environmental understanding, they primarily focus on online assistance during walking, rather than a contextual assessment of environmental risks along pedestrian routes. Furthermore, evaluations are often conducted in geographically limited settings, leaving their generalizability across diverse urban environments less explored. These limitations motivate approaches that leverage VLMs to systematically identify and quantify potential hazards, construct spatially grounded risk representations, and support offline pedestrian route assessment across geographically diverse environments.}

\new{Building on these capabilities, we propose a VQA-based framework for proactive urban risk assessment and assistive navigation (see Figure \ref{fig:pipeline}). First, we introduce a large-scale, geographically diverse dataset comprising more than 4,800 annotated urban images from 89 cities across all continents, sourced from Mapillary\footnote{\url{www.mapillary.com}}. Overall, the dataset includes over 95,000 answered questions addressing environmental features and potential pedestrian hazards, providing a broad basis for evaluating risk assessment across heterogeneous environments. To further evaluate the framework from a user perspective, we complement this dataset with 20 street sequences recorded in Barcelona using Biel Glasses smart glasses\footnote{\url{https://bielglasses.com}}. These sequences provide real-world urban imagery for a user study involving 20 participants, aimed at assessing the agreement between the model's risk estimations and users' perceived risk, as well as their trust in the model's assessments.}

\new{Second, we propose a three-level hierarchical VQA structure that moves beyond flat question-answering approaches to progressively characterize environmental elements and their associated risk. Level-1 questions identify relevant environmental features, while conditional follow-up questions at Levels 2 and 3 provide increasingly detailed descriptions and assess different aspects and levels of potential risk. For example, rather than simply detecting a pedestrian crossing, the framework can further examine its traffic conditions and signalization to obtain a more nuanced risk assessment. Each response contributes to a structured event with an associated risk score.}

Furthermore, we develop a weighted aggregation mechanism that combines these events into normalized street-segment risk values, which are represented through risk event maps with four risk bands: \new{\textit{Safe}, \textit{Low Risk}, \textit{Moderate}, and \textit{High Risk}}. This spatial representation provides quantitative information about potential hazards along pedestrian routes, supporting their proactive assessment before a journey. Finally, we benchmark five VQA architectures---ViLT \cite{vilt}, LLaVA \cite{llava}, InstructBLIP \cite{insblip}, Qwen-VL \cite{Qwen2-VL}, and \new{Gemma-4 \cite{gemma}}---to investigate their effectiveness in identifying and characterizing urban hazards and generating risk-aware representations.

The remainder of this paper is organized as follows. Section~\ref{sec:RW} reviews related work on assistive navigation technologies and VQA systems. \new{Section~\ref{sec:methodology} presents the proposed methodology, including the hierarchical query structure, the standardization of the VQA queries, the risk assessment formulation and the risk event map generation. Section~\ref{sec:experiments} describes the experiments: the dataset construction, the evaluation of hazard detection, hazard characterization and risk estimation, a user study assessing human agreement with the predicted risk bands, and a discussion of the limitations of this work.} Finally, Section~\ref{sec:conlcusions} concludes the paper by summarizing the main contributions and discussing their potential impact on assistive navigation technologies for \new{people with low vision}.

\section{Related Work}\label{sec:RW}

\subsection{Assistive Navigation}

Traditional pedestrian navigation systems have primarily focused on core perception tasks required for safe movement in urban environments, including real-time detection of surrounding pedestrians \cite{buck, cao, BRUNETTI201817}, walkable surface segmentation \cite{sidewalk}, and robust self-localization in GPS-denied or degraded settings \cite{ego,zhang,moisan2025mapin}. Even routine activities---such as determining whether it is safe to cross a street or identifying stairs---can pose substantial risks for people with low vision \cite{hwang2024safecrossinterpretablerisk, wangDeepLeaningbasedUltrafast2022}. However, these perception modules are typically developed and evaluated in isolation, rather than as tightly integrated components of end-to-end assistive navigation systems. Beyond core perception, several works have explored complementary modalities and representations to enrich navigation support. Augmented reality interfaces \cite{kumar} offer intuitive visual overlays for users with partial vision, while landmark-based systems \cite{hile, zhy} leverage salient environmental features to improve route following and spatial orientation. More recently, research has shifted toward personalized navigation by jointly modeling pavement infrastructure, walking environments, and individual pedestrian profiles to enable constraint-aware, preference-sensitive path planning \cite{shah_et_al, Fang02102015, novack}.

Despite these advances, existing methods struggle to cope with the inherent unpredictability of real-world urban environments. Dynamic obstructions---such as construction barriers, potholes, or unexpected crowds---are difficult to anticipate and represent using fixed perception pipelines. Moreover, existing systems lack a persistent, risk-aware spatial representation that can inform route-planning decisions. This motivates the use of a flexible, query-driven VQA model capable of reasoning about novel environmental conditions without task-specific retraining.

\subsection{Vision-Language Models}

Vision-Language Models (VLMs) represent a paradigm shift in how machines interpret visual content by bridging computer vision and NLP. CLIP \cite{pmlr-v139-radford21a} pioneered the use of contrastive learning to align image and text embeddings in a shared semantic space, enabling zero-shot classification and retrieval tasks. Building on this foundation, VQA models extend the interaction paradigm by generating free-form textual responses conditioned on both visual input and natural language queries.

Modern VQA architectures can be broadly categorized into discriminative and generative approaches. Discriminative models such as ViLT \cite{vilt} frame VQA as a classification task, predicting answers from a fixed vocabulary using vision-language transformers. In contrast, generative approaches such as BLIP-2 \cite{blip} and LLaVA \cite{llava}---representative of the broader family of Multimodal Large Language Models (MLLMs)---leverage autoregressive decoding to produce open-ended responses, offering greater flexibility across diverse question types.

The generalization capabilities of VQA models have enabled diverse applications beyond traditional question answering. Recent efforts have demonstrated their utility in open-vocabulary object detection \cite{kuo2023fvlmopenvocabularyobjectdetection,du2024lamidetropenvocabularydetectionlanguage,zhuu}, where natural language descriptions replace fixed category labels, and visual grounding \cite{Xiao_2024,zhao2023unleashingtexttoimagediffusionmodels}, which localizes objects or regions based on textual queries.

Latest work applies VLMs to blind and low-vision assistance \cite{merchant2024generatingcontextuallyrelevantnavigationinstructions, llava, zhao2024vialmsurveybenchmarkvisually}. For example, \cite{merchant2024generatingcontextuallyrelevantnavigationinstructions} demonstrated context-aware navigation guidance, while \cite{llava} fine-tuned VLMs for practical suggestions using LoRA. Applications such as Be My AI \cite{be_my_ai} adopt a reactive question-answering paradigm. WalkVLM \cite{yuanWalkVLMAidVisually2025} advances this further via hierarchical reasoning (perception–comprehension–decision) and temporal-aware prediction to reduce redundancy---demonstrating strong results on walking guidance tasks. However, WalkVLM is optimized for generating concise, reactive walking reminders from video streams, and does not produce a persistent spatial representation of the environment. As such, it cannot support proactive route planning or risk-aware path selection, which require reasoning over accumulated environmental information across a trajectory. 

\new{Complementing these efforts, \cite{morales} propose a VQA-based assistive navigation framework for low-vision users that aggregates VQA outputs into a persistent semantic event map georeferenced using GPS data, rather than generating per-frame walking instructions. Their unified, query-driven VQA backbone addresses diverse urban challenges---from detecting construction hazards to identifying traffic light states---without requiring task-specific vision pipelines. However, their evaluation is based on a custom dataset of over 1,300 annotated images collected exclusively in Barcelona. Our work builds upon this spatial reasoning paradigm while introducing three key extensions. First, we substantially broaden the geographical scope of the evaluation through a large-scale dataset covering 89 cities across all continents, enabling the assessment of generalization across diverse urban environments. Second, we introduce a richer three-level hierarchical query structure that progressively characterizes environmental features and their associated hazards through conditional questions. Third, we move beyond semantic event mapping by formulating a weighted risk assessment mechanism that aggregates detected events into quantitative, normalized risk scores for street segments, enabling the characterization and comparison of potential risks along pedestrian routes. These extensions provide a more comprehensive framework for geographically generalizable urban risk assessment and risk-aware assistive navigation.}

\subsection{Assistive Datasets}

Progress in assistive navigation depends on representative datasets. Accessibility-focused datasets such as VizWiz \cite{gurari2018vizwizgrandchallengeanswering, vizwiz} collect real-world questions from blind users, yet are largely object-centric rather than navigation-driven. Other semantic datasets (\textit{e.g.}, VIALM \cite{zhao2024vialmsurveybenchmarkvisually} and small-scale user studies \cite{merchant2024generatingcontextuallyrelevantnavigationinstructions}) remain limited in size and diversity.

Recent large-scale efforts improve coverage and standardization. The Walking Awareness Dataset (WAD) \cite{yuanWalkVLMAidVisually2025} contains 12,000 video-annotation pairs across multiple regions, but is designed primarily to train and evaluate models for generating walking reminders, rather than to support spatial risk assessment. GUIDE DOG \cite{kimGuideDogRealWorldEgocentric2025} introduces 22,000 pedestrian-view image–description pairs with a human-AI collaborative annotation pipeline for fine-grained perception evaluation. However, GUIDE DOG annotations follow a guidance generation paradigm---describing surroundings and obstacles to a user in motion---and do not provide the hazard-level labels or georeferenced spatial structure required for route-level risk mapping. Neither dataset supports the aggregation of scene-level VQA outputs into navigable risk event maps, which is the objective of our work. Our dataset is therefore designed from the ground up around a hierarchical multicategory query structure targeting pedestrian safety categories, with annotations structured to feed directly into a weighted risk scoring and risk event map generation pipeline, validated across geographically diverse urban environments.

\subsection{Scene Representation}

Scene graphs provide structured, semantic representations of environments by encoding objects as nodes and their relationships as edges \cite{Rosinol20rss-dsg}. This abstraction enables efficient spatial reasoning and supports natural language interfaces for scene querying. Originally developed for indoor robotic applications---where controlled environments and static infrastructure make graph construction tractable---scene graphs have been employed for semantic guidance \cite{dai2024optimalscenegraphplanning}, object search and task planning \cite{tasko}, and viewpoint-aware scene understanding \cite{liu2023birdseyeviewscenegraphvisionlanguage}.

The dynamic nature of real-world environments---where objects like furniture or personal items may relocate---has motivated extensions to scene graphs that model temporal object persistence \cite{Zhou_Liu_Zhao_Liang_2023}. Recent work has further integrated scene graphs with LLMs such as GPT-3.5, enabling natural language queries over structured spatial data and allowing agents to dynamically retrieve context-relevant information encoded in graph nodes \cite{Werby-RSS-24}. However, despite these advances, scene graph construction remains largely confined to indoor settings, where the bounded spatial extent and predictable structure of environments simplify node definition and edge inference. Outdoor urban environments, by contrast, present open-ended, constantly changing scenes that make exhaustive graph construction prohibitively expensive \new{to build and maintain at city scale}.

While scene graphs offer rich semantic structure, many navigation applications require simpler, more computationally efficient representations. Event maps---a lightweight alternative---maintain spatial information without the full complexity of object-relationship graphs \cite{morales}. By encoding discrete events (\textit{e.g}., detected hazards, crosswalks, obstacles) at specific geographic coordinates, event maps provide a pragmatic middle ground between detailed scene graphs and raw image streams. \new{This representation is particularly well-suited for offline route assessment: risk is computed once from previously captured imagery, stored with its location, and reused by any number of users and routes.}

Our approach adopts the event map paradigm as its spatial backbone, extending it with multi-level, VLM-derived risk scores rather than simple binary hazard flags. This produces semantically enriched, georeferenced risk event maps that are both computationally tractable\new{, since the analysis is performed offline and only once per location,} and expressive enough to support meaningful route selection for \new{people with low vision}.
\section{Method}\label{sec:methodology}

\noindent \new{We propose a model for generating risk event maps to support assistive urban navigation for people with low vision. Inspired by the event map paradigm of \cite{morales}, our approach uses georeferenced street-level imagery to identify and characterize potential pedestrian hazards. A three-level hierarchical query structure guides a VQA model in progressively detecting hazards and refining their severity based on contextual information. The resulting predictions are combined through a weighted noisy-OR formulation to estimate an image-level risk score, which is classified into four risk bands (\textit{Safe}, \textit{Low Risk}, \textit{Moderate}, and \textit{High Risk}). These georeferenced image-level risk estimates are subsequently aggregated over individual street segments along a given route, producing risk event maps that provide people with low vision with a localized assessment of pedestrian safety.}

\begin{figure*}[t!]  
    \centering
    \includegraphics[width=1\linewidth]{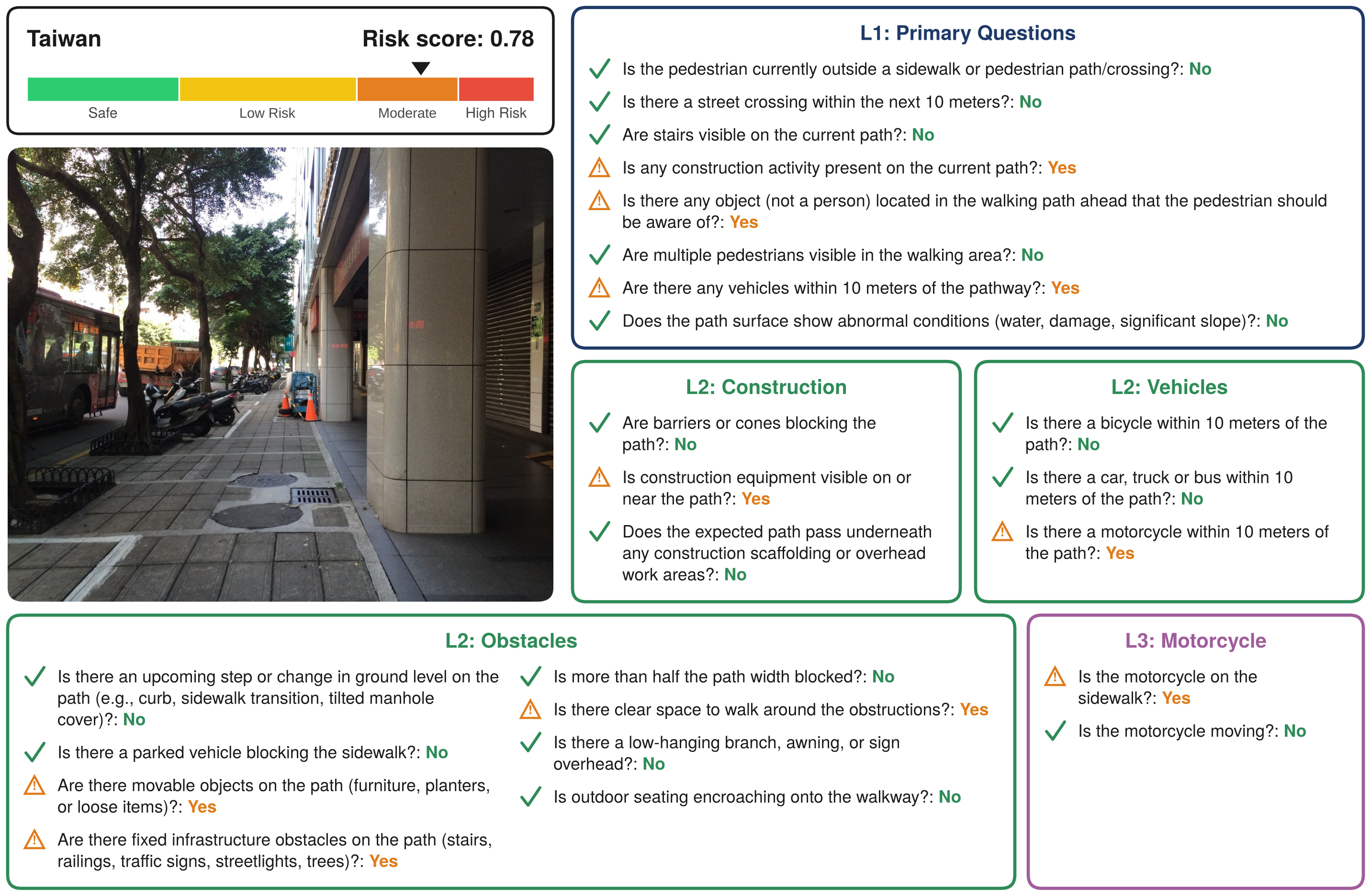}
    \caption{\new{Example of the hierarchical multicategory query structure applied to a street scene in Taiwan. The hierarchy begins with Level-1 (Primary Questions)---blue---to detect the presence of broad hazard categories. Affirmative detections trigger conditional follow-up queries at Level-2 (Detailed Characterization)---green---and Level-3 (Fine-grained Refinement)---purple---, ensuring high semantic density where hazards exist. Potential hazards are marked with orange warning signs, while safe conditions are indicated with green check marks. The Level-2 and Level-3 answers refine the severity of the hazard they elaborate on, and the resulting hazard strengths are combined into a single normalized risk score ($0.78$ in this example, indicated by the black marker on the bar and falling in the \textit{Moderate} band), which serves as the foundational data point for generating risk event maps.}}
    \label{fig:qa}
\end{figure*}

\subsection{Hierarchical Query Structure Design} \label{sec:mcquest}

To obtain comprehensive and fine-grained scene descriptions, we designed a three-level hierarchical query structure, illustrated in Figure \ref{fig:qa}. At Level 1, coarse-grained categorical questions identify potential hazards and trigger conditional follow-up queries at Levels 2 and 3, which progressively refine their characterization and provide contextual evidence for subsequent risk estimation. This conditional structure avoids redundant questions and allocates descriptive detail only to hazards identified at the preceding levels.  

\textbf{Level-1} questions establish the main scene categories relevant to pedestrian safety for people with low vision. Each question focuses on a distinct hazard class: (i)~\textit{Stairs} — whether steps or changes in ground level are visible on the current path; (ii)~\textit{Crossings} — whether a street crossing exists within approximately 10 meters; (iii)~\textit{Construction} — whether construction activity or pathway detours are visible; (iv)~\textit{Obstacles} — whether any object in the walking path ahead requires the pedestrian's attention; (v)~\textit{Crowding} — whether multiple pedestrians are visible in the walking area; (vi)~\textit{Vehicles} — whether any vehicle is within 10 meters of the path; (vii)~\textit{Surface} — whether the path surface shows abnormal conditions such as wetness, damage, or significant slope; and (viii)~\textit{Non-Sidewalk} — whether the visible path lies outside a designated sidewalk or pedestrian crossing. A positive response to any Level-1 question unlocks a targeted set of Level-2 follow-up queries specific to that category.

\textbf{Levels 2 and 3} provide an increasingly detailed characterization of any detected element. Level-2 questions provide a more detailed characterization of the detected hazard---for instance, distinguishing whether a detected vehicle is a bicycle, car, or motorcycle, or determining whether a crossing is signalized and currently occupied by traffic. Where Level-2 confirms a specific condition, Level-3 refines it further: a detected vehicle in a crossing lane triggers a question about whether it is moving or stationary, and a bicycle triggers queries about its direction of travel and whether it occupies a dedicated bike lane. Note that not all hazard categories require Level-3 queries; deeper questioning is reserved for those cases where additional granularity has direct safety implications. This cascading structure ensures that hazard descriptions are both precise and contextually grounded, while keeping the total number of questions minimal.  \new{In total, the hierarchy comprises 44 questions: 8 at Level 1, 27 at Level 2 and 9 at Level 3, with up to 9 follow-ups for a single hazard (\textit{Vehicles}) and as few as 2 for others (\textit{Crowding}, \textit{Non-Sidewalk}).}

\subsection{VQA Query Standardization}

To ensure consistent and safe responses, all models receive a shared base context prepended to each question. This context defines the model's role and guides its responses:

\begin{center}
    \textit{``You are an expert at detecting pedestrian 
    obstacles for people with low vision. Answer only 
    with Yes or No.''}
\end{center}

By applying this base context, we ensure that all outputs are constrained to a concise, actionable binary format. This uniformity is critical for direct performance comparison against the manually annotated ground truth. Evaluation metrics are detailed in Section~\ref{sec:ef}.

\subsection{Image- and Street-Level Risk Assessment}\label{sec:riskform}

To quantify pedestrian risk from the hierarchical VQA predictions, we adopt a two-stage risk assessment formulation. First, the answers associated with each image are integrated into an image-level risk score that accounts for both the presence and severity of detected hazards. Second, these scores are spatially aggregated across street segments to estimate segment-level risk and generate the corresponding risk event maps. This formulation provides a unified representation of pedestrian risk at both the image and street-segment levels.

Let $S$ be a segment of the traveled route, and $M_S$ be the set of nodes contained within that segment, where each node is an analyzed image together with its GPS position. The total risk for the segment, $R_{seg}$, is defined as the maximum risk score found among its nodes:

\begin{equation} \label{eq:1}
    R_{seg} = \max_{m \in M_S} (R_{img}(m))
\end{equation}

This max-aggregation strategy follows a conservative safety principle, in which the overall risk of a segment is determined by its most hazardous point. This approach ensures that even a single dangerous location (such as a construction zone or a highly sloped sidewalk) elevates the entire segment's risk classification, preventing users from being routed through partially unsafe areas.

\new{Following this definition, risk values are grouped into four bands: \textit{Safe} ($R\le0.3$), \textit{Low Risk} ($0.3<R<0.65$), \textit{Moderate} ($0.65\le R<0.85$) and \textit{High Risk} ($R\ge0.85$).}

\new{The risk score of each image, $R_{img}(m)$, is obtained by combining the Level-1 hazards, weighted by their severity and refined by the Level-2 and Level-3 answers:}

\begin{equation}\label{eq:2}
    R_{img}(m) = 1 - \prod_{i \in Q} \left( 1 - w_i \cdot y_i \cdot s_i \right)
\end{equation}

\new{\textbf{Where:}}

\new{$Q$ is the set of primary (Level-1) questions evaluated for the image, which serve as the foundation for risk assessment.}
    
\new{$w_i$ is the weight determined by the hazard tier $T$ of question $i$, reflecting the severity and immediacy of different hazard categories:
\begin{equation}
    w_i = 
    \begin{cases} 
        0.85 & \text{if } i \in T_{\text{critical}} \text{ \small \begin{tabular}{@{}l@{}}\textit{Construction}, \textit{Surface}, \\ \textit{Non-Sidewalk}\end{tabular}} \\
        0.6 & \text{if } i \in T_{\text{high}} \text{ \textit{Crossings}, \textit{Stairs}, \textit{Obstacles}} \\
        0.3 & \text{if } i \in T_{\text{low}} \text{ \textit{Crowding}, \textit{Vehicles}}
    \end{cases}
\end{equation}
}

\new{Critical hazards ($w_i = 0.85$) represent immediate threats to safe navigation, such as construction zones that may block the entire sidewalk, hazardous surface conditions like broken pavement or ice, and pedestrians walking in the roadway. High-priority hazards ($w_i = 0.6$) include features requiring careful navigation such as crossings, stairways, and path-blocking obstacles. Low-priority hazards ($w_i = 0.3$) encompass dynamic elements like nearby pedestrians and vehicles that require awareness but do not directly impede travel. These weights are set relative to the band limits above: a single low-priority hazard never exceeds \textit{Safe}, a single high-priority hazard reaches at most \textit{Low Risk}, and \textit{Moderate} or \textit{High Risk} require a critical hazard or a combination of hazards.}
    
\new{$y_i\in\{0,1\}$ is the answer indicator of question $i$ in image $m$: $y_i=1$ if the answer is `Yes' (Hazard), and $y_i=0$ if it is `No' or unanswered.}

\new{$s_i\in[0,1]$ is the severity of hazard $i$ when it is present, computed recursively over the hierarchy so that Level-2 and Level-3 answers contribute to the score. For any question $n$ answered `Yes' (at any level):
    
    \begin{equation}\label{eq:strength}
        s_n = \Big[\sigma_n + (1-\sigma_n)\,A_n\Big]\,(1-M_n)
    \end{equation}
    \begin{equation}\label{eq:aggravating}
        A_n = 1-\!\!\prod_{c\in\mathcal{C}^{+}(n)}\!\!(1-y_c\,s_c)
    \end{equation}
    \begin{equation}\label{eq:mitigating}
        M_n = 1-\!\!\prod_{c\in\mathcal{C}^{-}(n)}\!\!(1-y_c\,s_c)
    \end{equation}
}

\new{where $\mathcal{C}^{+}(n)$ and $\mathcal{C}^{-}(n)$ are the follow-up questions of $n$ (Level-2 for a Level-1 question, Level-3 for a Level-2 question) that aggravate or mitigate it, respectively. $A_n$ and $M_n$ measure how much of this evidence has been confirmed with a `Yes': the more aggravating answers, the closer $A_n$ is to 1, which pushes $s_n$ towards 1, while mitigating answers increase $M_n$ and pull $s_n$ towards 0. $\sigma_n\in[0,1]$ is the starting severity of question $n$: $\sigma_i=0.5$ for Level-1 questions, and a question-specific value for Level-2 and Level-3 questions that sets how strongly each one modifies its parent. Without any answered follow-up, $s_n=\sigma_n$. The values of $\sigma_c$ were obtained through experimentation.}

\new{For instance, a lone surface hazard yields $R_{img}\approx0.43$ (\textit{Low Risk}), rising to $0.78$ (\textit{Moderate}) if both water or ice and damaged pavement are confirmed at Level-2. Similarly, a crossing with a green pedestrian signal and painted markings yields $0.08$ (\textit{Safe}), versus $0.53$ (\textit{Low Risk}) with no signal and moving vehicles, and $0.74$ (\textit{Moderate}) if an obstacle also blocks half of the path.}

\subsection{Risk Event Map Generation}\label{sec:eventmapg}

\new{To generate spatial representations of pedestrian risk, the image-level risk score of each node is aggregated over the street network according to its GPS position. The risk event maps are generated offline from previously captured street-level imagery (\textit{e.g.}, Mapillary or previously recorded walks), requiring no inference while the user is walking. Each node is assigned to the nearest street segment of the OpenStreetMap (OSM) graph, using segments of 40~m. When multiple nodes fall within the same segment, the maximum image-level risk score is retained (Equation~\ref{eq:1}), ensuring that a single high-risk node raises the risk classification of the entire segment. Finally, each segment is color-coded according to its aggregated risk band and overlaid on the GPS trajectory, producing a risk event map that can be used to assess and compare alternative routes.}

\section{Experiments} \label{sec:experiments}

\noindent We conduct an extensive evaluation of the proposed framework on a newly constructed dataset comprising urban scenes from 89 cities worldwide. We first benchmark five VQA models---ViLT, LLaVA, InstructBLIP, Qwen-VL, and Gemma-4---for Level-1 hazard detection (Section~\ref{sec:ef}). We then assess their ability to characterize detected hazards through the hierarchical Level-2 and Level-3 queries (Section~\ref{sec:charac}), and evaluate the resulting risk estimates at both image and street-segment levels (Section~\ref{sec:riskeval}). Finally, the user study (Section~\ref{sec:uc}) examines the agreement between model-predicted risk bands and the risk perceptions of 20 participants, as well as their trust in the model's assessments.

\subsection{Global Multi-City Dataset Construction}

\begin{figure}[t!]
    \centering
    \includegraphics[width=1\linewidth]{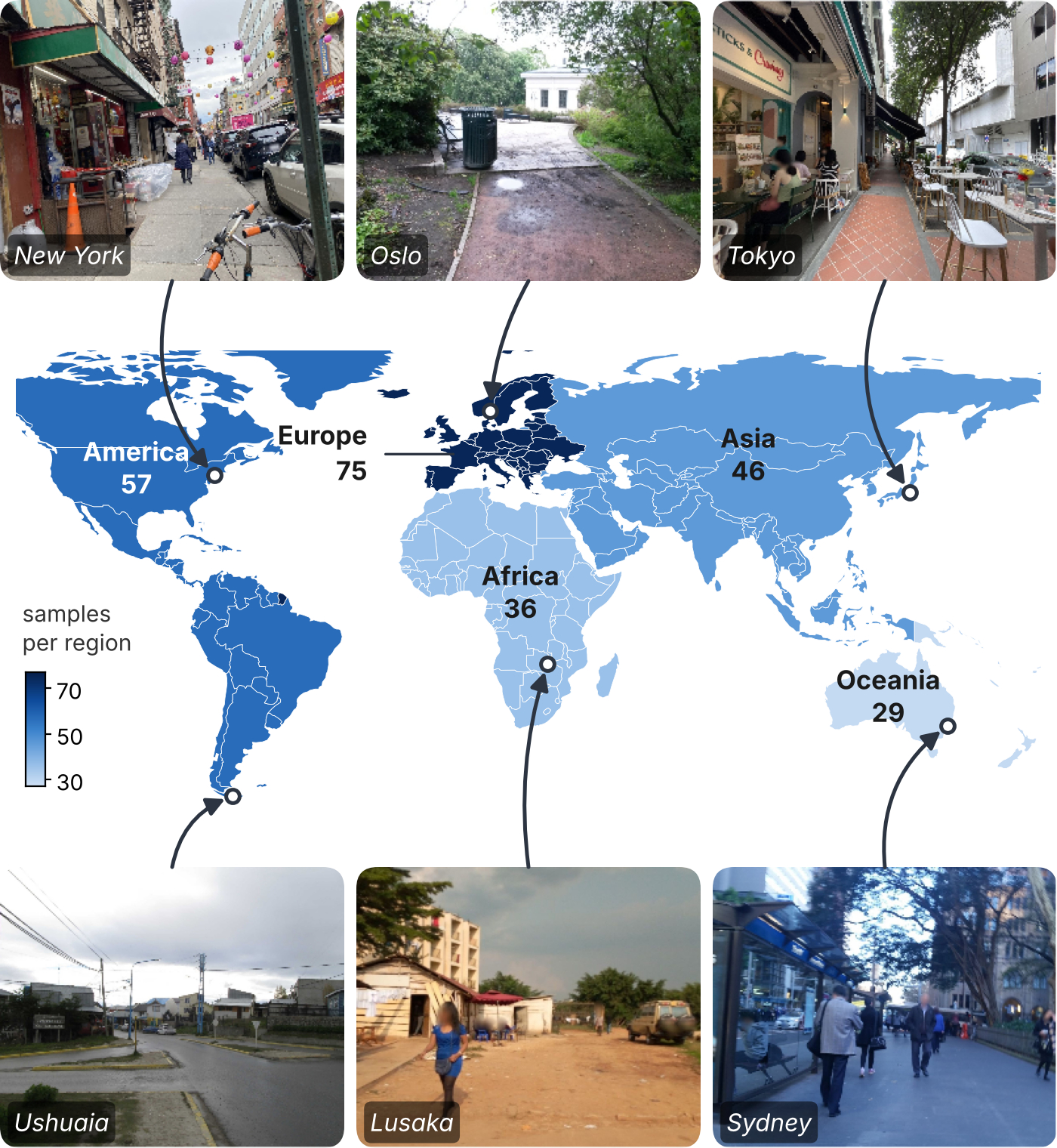}
    \caption{Geographic distribution of the dataset across \new{89 cities}. \new{Continents are shaded by the number of sequences they contribute (darker = more)}. Representative street-view images illustrate the diversity of urban environments.}
    \label{fig:dataset_map}
\end{figure}

The dataset is designed to promote and assess the generalization capabilities of the proposed method across diverse urban contexts worldwide. By including a geographically broad range of environments, our goal is to evaluate its applicability beyond specific cities or regions, encompassing different urban layouts, infrastructures, and pedestrian environments. To this end, we collected image sequences from \new{89} cities across all continents, covering a wide range of environments, from dense metropolitan downtown areas to unpaved roads in rural regions (see Figure \ref{fig:dataset_map}). The resulting dataset comprises \new{244} sequences.

The images were obtained from two sources: (i) the Mapillary platform and (ii) self-recorded sequences captured in Barcelona using Biel Glasses smart glasses.

To generate the ground truth (GT), all annotations were manually performed by human annotators. From each sequence, 20 images were uniformly sampled and annotated using the hierarchical multicategory query structure described in Section \ref{sec:mcquest}, resulting in over \new{4,800} labeled images, associated with a total of over \new{95,000} answered questions.

\subsection{Hazard Detection}\label{sec:ef}

The performance of each model is quantitatively assessed by comparing its automated responses against the manually annotated GT. \new{We first evaluate hazard detection, that is, the Level-1 answers on all images and all eight primary questions. Because the query structure is hierarchical, the Level-2 and Level-3 answers (Section~\ref{sec:charac}) and the resulting risk score (Section~\ref{sec:riskeval}) are evaluated separately and never pooled with them.} We employ a comprehensive suite of binary classification metrics to evaluate the reliability of the system in hazard-sensitive urban scenarios:

\begin{itemize}
    \item \textbf{Accuracy:} The proportion of total predictions (both positive and negative) that were correct, providing an overall measure of model reliability.
    \item \textbf{Precision:} The ability of the model to avoid false alarms, calculated as the ratio of true positives to all positive predictions. This metric is crucial for preventing user alert fatigue.
    \item \textbf{Recall:} A critical safety metric representing the model's ability to detect all existing hazards (\textit{e.g.}, ensuring no stairs or vehicles are missed). High recall is paramount in assistive navigation where undetected hazards pose direct physical risks.
    \item \textbf{Specificity:} The model's ability to correctly identify the absence of a hazard, which is essential to avoid over-alerting the user and maintaining system trust.
    \item \textbf{F1-score:} The harmonic mean of precision and recall, providing a single balanced metric for model comparison that accounts for both false positives and false negatives.
    \item \new{\textbf{Balanced accuracy:} The mean of recall and specificity. Hazard prevalence varies widely across categories, so this metric is not dominated by the most frequent answer.}
\end{itemize}

\new{Hazard prevalence, \textit{i.e.}, the share of images in which a hazard is present, ranges from about 2\% to 81\% of the images depending on the category. To put the metrics in context, we therefore also report the prevalence of each hazard and the scores of trivial always-``Yes'' and always-``No'' predictors.}

A significant challenge in evaluating hierarchical VQA is the conditional nature of the questions: Level-2 and Level-3 follow-ups are only annotated when the annotators detected the Level-1 hazard, so no GT exists for them otherwise. The Level-2 and Level-3 metrics are therefore computed exclusively on the questions with a corresponding GT answer, \new{each posed to the model regardless of its Level-1 answer}. This ensures that we measure the model's ability to describe a hazard only when it is actually present, preventing artificial inflation of error rates due to cascading misclassifications.

\begin{table}[h!]
    \centering
    \caption{\new{Level-1 hazard detection of all VQA models (positive-answer prevalence 0.271). The last two rows are trivial predictors.}}
    \label{tab:model_performance}
    \new{\begin{tabular}{lcccccc}
        \toprule
        \textbf{Model} & \textbf{Acc.} & \textbf{F1} & \textbf{Prec.} & \textbf{Rec.} & \textbf{Spec.} & \textbf{Bal. Acc.} \\
        \midrule
        ViLT         & 0.532 & 0.393 & 0.303 & 0.559 & 0.521 & 0.540 \\
        LLaVA        & 0.744 & 0.461 & 0.538 & 0.403 & 0.871 & 0.637 \\
        InstructBLIP & \textbf{0.841} & 0.656 & \textbf{0.797} & 0.558 & \textbf{0.947} & 0.752 \\
        Qwen-VL      & 0.780 & 0.641 & 0.575 & 0.725 & 0.800 & 0.763 \\
        Gemma-4      & 0.823 & \textbf{0.712} & 0.638 & \textbf{0.805} & 0.830 & \textbf{0.818} \\
        \midrule
        Always ``Yes'' & 0.271 & 0.426 & 0.271 & 1.000 & 0.000 & 0.500 \\
        Always ``No''  & 0.729 & 0.000 & --    & 0.000 & 1.000 & 0.500 \\
        \bottomrule
    \end{tabular}}
\end{table}

As shown in Table~\ref{tab:model_performance}, \new{Gemma-4 and Qwen-VL are the two strongest models and the ones with the most balanced detection profiles. Gemma-4 reaches the highest F1-score (0.712), recall (0.805) and balanced accuracy (0.818), while Qwen-VL follows closely (F1-score 0.641, recall 0.725, balanced accuracy 0.763).} This balanced profile is critical for assistive navigation, where both hazard detection and user trust are essential. \new{Since 73\% of the answers are negative, the always-``No'' predictor already reaches an accuracy of 0.729, so we rely on recall, F1-score and balanced accuracy, which are more informative than accuracy in this setting.}

\new{ViLT performs below the always-``No'' predictor (accuracy 0.532) and cannot discriminate hazards reliably: with a specificity of only 0.521 and a precision of 0.303, it raises a false alarm in nearly half of the hazard-free cases.} Such behavior would overwhelm users with false alarms, making the system unusable. In contrast, InstructBLIP's low recall (\new{0.558}) represents an equally critical but opposite failure---missing \new{more than 40\%} of actual hazards poses unacceptable safety risks for users with low vision. LLaVA achieves moderate performance on all metrics, but still fails to detect \new{about 60\% (recall 0.403)} of the hazards present.

\new{Gemma-4 and Qwen-VL offer complementary profiles, and we therefore use both in the rest of the evaluation: Gemma-4 provides the best pooled detection, with fewer false alarms than Qwen-VL, whereas Qwen-VL is the most sensitive model on the safety-critical hazard categories (Table~\ref{tab:topic_performance}), where missing a hazard is most costly.}

\begin{table}[h!]
    \centering
    \caption{\new{Level-1 detection by hazard category for Qwen-VL and Gemma-4. Prev.\ is the share of images in which the hazard is present.}}
    \label{tab:topic_performance}
    \new{\footnotesize\setlength{\tabcolsep}{2.5pt}
    \begin{tabular}{lc cccc cccc}
        \toprule
         & & \multicolumn{4}{c}{\textbf{Qwen-VL}} & \multicolumn{4}{c}{\textbf{Gemma-4}} \\
        \cmidrule(lr){3-6}\cmidrule(lr){7-10}
        \textbf{Hazard} & \textbf{Prev.} & \textbf{Rec.} & \textbf{Spec.} & \textbf{F1} & \textbf{BA} & \textbf{Rec.} & \textbf{Spec.} & \textbf{F1} & \textbf{BA} \\
        \midrule
        Construction & 0.021 & \textbf{0.920} & \textbf{0.938} & \textbf{0.378} & \textbf{0.929} & 0.770 & 0.922 & 0.279 & 0.846 \\
        Crossings    & 0.133 & \textbf{0.861} & 0.834 & 0.585 & \textbf{0.848} & 0.539 & \textbf{0.968} & \textbf{0.617} & 0.754 \\
        Obstacles    & 0.808 & 0.533 & \textbf{0.701} & 0.664 & 0.617 & \textbf{0.781} & 0.463 & \textbf{0.818} & \textbf{0.622} \\
        Non-Sidewalk & 0.187 & \textbf{0.916} & 0.264 & 0.358 & 0.590 & 0.543 & \textbf{0.839} & \textbf{0.485} & \textbf{0.691} \\
        Crowding     & 0.336 & 0.900 & \textbf{0.830} & \textbf{0.805} & \textbf{0.865} & \textbf{0.980} & 0.587 & 0.701 & 0.783 \\
        Stairs       & 0.041 & \textbf{0.905} & 0.956 & \textbf{0.617} & \textbf{0.931} & 0.741 & \textbf{0.963} & 0.571 & 0.852 \\
        Surface      & 0.125 & 0.498 & \textbf{0.939} & \textbf{0.518} & 0.719 & \textbf{0.633} & 0.845 & 0.465 & \textbf{0.739} \\
        Vehicles     & 0.521 & 0.839 & \textbf{0.799} & \textbf{0.829} & \textbf{0.819} & \textbf{0.941} & 0.570 & 0.805 & 0.755 \\
        \bottomrule
    \end{tabular}}
\end{table}

Table~\ref{tab:topic_performance} \new{details the detection of each hazard category. Qwen-VL is highly sensitive to the safety-critical categories: its recall reaches 0.92 for construction zones, 0.91 for stairs and 0.92 for non-sidewalk paths, the highest among all evaluated models (the second best being 0.77, 0.78 and 0.68, respectively). The price of this sensitivity is a lower precision on the rarest hazards and a specificity of 0.26 for non-sidewalk paths: the model prefers raising a false alert to missing a hazard, a behavior that is desirable in assistive navigation and that the risk score must absorb (Section~\ref{sec:riskeval}). Gemma-4 follows the opposite strategy, being more conservative: it has the higher specificity on crossings (0.97) and non-sidewalk paths (0.84), the higher F1-score on obstacles (0.82) and non-sidewalk paths (0.48), and the higher recall on surface hazards (0.63), obstacles (0.78) and crowding (0.98), at the cost of missing more construction zones, crossings and stairs. Averaged over the eight categories, both models obtain the same F1-score (0.59), while Qwen-VL has the higher mean recall (0.80 \textit{vs.} 0.74). Both models detect vehicles and crowding with high recall, although Gemma-4 raises 1,338 false alarms for crowding, and both struggle with surface hazards (subtle visual cues) and especially with obstacles (present in 81\% of the images and very heterogeneous), their most frequently missed hazard (1,839 and 864 cases for Qwen-VL and Gemma-4); improving their recall, \textit{e.g.}, through class-balanced fine-tuning, is a natural next step. Errors are slightly lower on our Barcelona recordings than on Mapillary images (error rate 0.14 \textit{vs.} 0.18 for Gemma-4 and 0.21 \textit{vs.} 0.22 for Qwen-VL), and no continent stands out (0.16--0.25).}

\subsection{Hazard Characterization}\label{sec:charac}

\new{We now evaluate the Level-2 and Level-3 follow-up questions, which describe a hazard once it has been detected (\textit{e.g.}, whether a crossing is signalized). Following the protocol of Section~\ref{sec:ef}, each question is evaluated only where the annotators triggered it, which yields 50,112 Level-2 and 6,741 Level-3 answers over the 36 follow-up questions of Section~\ref{sec:mcquest}}.

\begin{table}[h!]
    \centering
    \caption{\new{Characterization (Level-2 and Level-3 questions). The ``Always No'' rows are the trivial predictor; BA is the balanced accuracy.}}
    \label{tab:characterization}
    \new{\begin{tabular}{lcccccc}
        \toprule
        \textbf{Model} & \textbf{Acc.} & \textbf{F1} & \textbf{Prec.} & \textbf{Rec.} & \textbf{Spec.} & \textbf{BA} \\
        \midrule
        \multicolumn{7}{l}{\textit{Level 2 ($n=50{,}112$, prevalence 0.376)}} \\
        Qwen-VL        & \textbf{0.735} & \textbf{0.665} & \textbf{0.634} & 0.699 & \textbf{0.756} & \textbf{0.728} \\
        Gemma-4        & 0.675 & 0.645 & 0.547 & \textbf{0.785} & 0.608 & 0.697 \\
        Always ``No''  & 0.624 & 0.000 & --    & 0.000 & 1.000 & 0.500 \\
        \midrule
        \multicolumn{7}{l}{\textit{Level 3 ($n=6{,}741$, prevalence 0.265)}} \\
        Qwen-VL        & \textbf{0.759} & \textbf{0.542} & \textbf{0.546} & \textbf{0.537} & \textbf{0.839} & \textbf{0.688} \\
        Gemma-4        & 0.668 & 0.422 & 0.392 & 0.458 & 0.744 & 0.601 \\
        Always ``No''  & 0.735 & 0.000 & --    & 0.000 & 1.000 & 0.500 \\
        \bottomrule
    \end{tabular}}
\end{table}
 
\new{Qwen-VL is the stronger characterizer (Table~\ref{tab:characterization}): its balanced accuracy is 0.73 at Level 2 and 0.69 at Level 3, against 0.70 and 0.60 for Gemma-4. It is the only model above the always-``No'' accuracy at Level 3, and it has the higher balanced accuracy on 23 of the 36 questions. Gemma-4 is more permissive, with the higher Level-2 recall (0.785 \textit{vs.} 0.699), and is better on some questions, such as \textit{Water/Ice} (0.84 \textit{vs.} 0.68).}

\new{Both models do well on questions with clear visual categories, such as \textit{On Unpaved} (0.92 for Qwen-VL) or \textit{Motorcycle} (0.90). Their weakest questions fall into two groups: those whose answer is almost constant, such as \textit{On Road} (a pedestrian on the roadway, positive in 94\% of the cases; balanced accuracy 0.40 and 0.42), where the models cannot separate the two classes, and those that need motion cues, such as whether a vehicle at a crossing is moving (0.50 for Qwen-VL), which cannot be answered reliably from a single image.}
 
\subsection{Risk Evaluation}\label{sec:riskeval}

\begin{figure}[t!]
    \centering
    \includegraphics[width=1\linewidth]{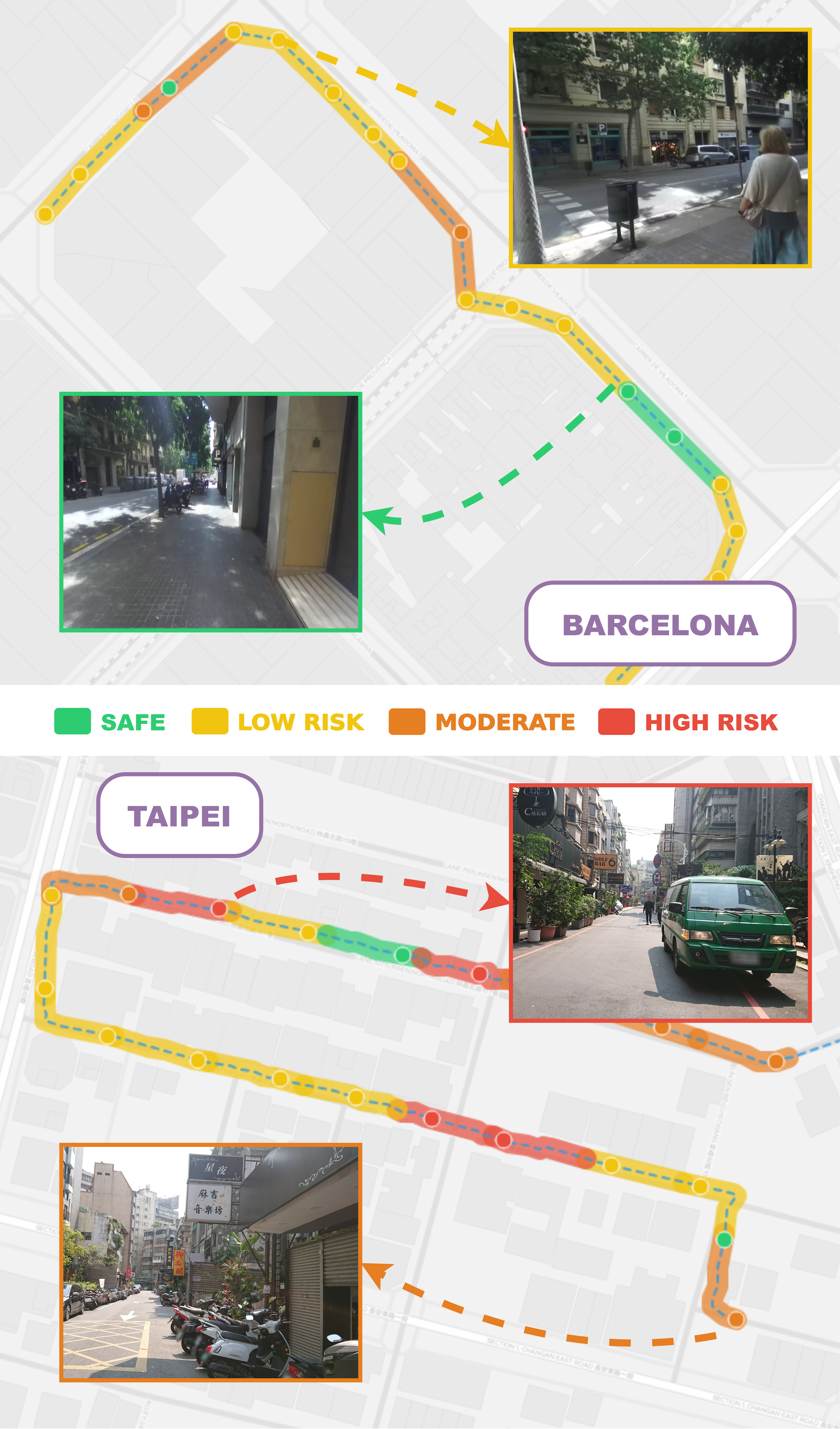}
    
    \caption{Risk event maps generated using Qwen-VL for routes in Barcelona (top) and \new{Taipei} (bottom). Street segments are color-coded into four risk bands---\textit{Safe} (green), \textit{Low Risk} (yellow), \textit{Moderate} (orange), and \textit{High Risk} (red)---based on the aggregated weighted risk scores derived from hierarchical VQA responses. \new{Gray segments have no associated image data and the dashed blue line is the GPS trajectory.} Circular markers denote node capture locations\new{, and dashed arrows link representative street-level images to their corresponding route locations}. \new{The images} illustrate the visual conditions \new{associated with} each risk band: a clear, unobstructed sidewalk (\textit{Safe}), \new{the approach to} a pedestrian crossing (\textit{Low Risk}), \new{a narrow street with densely parked scooters and limited visibility} (\textit{Moderate}), and \new{a narrow street with no separated pedestrian space, where a large vehicle shares the roadway with pedestrians} (\textit{High Risk}).}
    \label{fig:combined_event_maps}
\end{figure}

\new{Finally, we evaluate the risk estimates obtained by combining the hierarchical VQA predictions (Section~\ref{sec:riskform}). Figure~\ref{fig:combined_event_maps} illustrates the resulting risk event maps generated using Qwen-VL, following the procedure described in Section~\ref{sec:eventmapg}, for routes in Barcelona (top) and Taipei (bottom). These examples illustrate how image-level hazard assessments are translated into spatially aggregated route-level representations of pedestrian risk across geographically diverse urban environments.}

To quantitatively evaluate the precision of these automated risk assessments, we define the Risk Mean Absolute Error ($\text{MAE}_{\mathcal{R}}$), which measures the average deviation between model-generated risk scores ($R^{pred}$) and ground truth scores ($R^{gt}$) derived from human annotations across $N$ images or segments:
 
\begin{equation}
    \text{MAE}_{\mathcal{R}} = \frac{1}{N} \sum_{m=1}^{N} |R^{gt}(m) - R^{pred}(m)|
\end{equation}
 
\new{We complement the Risk MAE with the \textit{Critical Miss Rate} (CMR), defined as the proportion of ground-truth (GT) \textit{Moderate} or \textit{High Risk} cases that are incorrectly classified as \textit{Safe} or \textit{Low Risk}. This metric captures dangerous locations that the model fails to flag. Both metrics are evaluated at two spatial granularities: per image (4,874 images) and per 40~m street segment ($R_{seg}$; 2,968 segments across the 219 sequences with GPS metadata). For both evaluations, risk scores are computed using each model's own predictions across all hierarchy levels, with Level-2 and Level-3 follow-up questions triggered only when the model confirms the hazard at the preceding level.}
 
\begin{table}[t]
    \centering
    \caption{\new{Risk score per image and per 40~m segment, and mean processing time per image (single RTX 4090). Params B in billions. $\text{MAE}_{\mathcal{R}}$: mean absolute error; CMR: critical miss rate.}}
    \label{tab:risk_performance}
    \new{\footnotesize
    \setlength{\tabcolsep}{4pt}
    \begin{tabular}{lccccc}
        \toprule
        & \multicolumn{2}{c}{\textbf{$\text{MAE}_{\mathcal{R}}$}} & \multicolumn{2}{c}{\textbf{CMR}} & \\
        \cmidrule(lr){2-3}\cmidrule(lr){4-5}
        \textbf{Model (B params)} & \textbf{Image} & \textbf{Segment} & \textbf{Image} & \textbf{Segment} & \textbf{Time (s)} \\
        \midrule
        ViLT (0.12)         & 0.442 & 0.406 & 0.021 & 0.011 & 0.23 \\
        LLaVA (7.6)         & 0.287 & 0.299 & 0.738 & 0.693 & 7.1 \\
        InstructBLIP (4.0)  & 0.212 & 0.213 & 0.660 & 0.604 & 0.72 \\
        Qwen-VL (8.3)       & 0.254 & 0.246 & 0.286 & 0.214 & 23.7 \\
        Gemma-4 (8.0)       & 0.233 & 0.232 & 0.308 & 0.232 & 2.5 \\
        \bottomrule
    \end{tabular}}
\end{table}

\new{Table~\ref{tab:risk_performance} shows that error and critical misses must be read together. InstructBLIP attains the lowest $\text{MAE}_{\mathcal{R}}$ (0.21) but overlooks 66\% of the critical images and 60\% of the critical segments, and LLaVA 74\% and 69\%, in line with their low recall in detection. ViLT shows the opposite behavior: it almost never misses a critical location (CMR 0.02) because it flags hazards almost everywhere, which yields the largest error (0.44). Gemma-4 and Qwen-VL offer the best compromise, with an $\text{MAE}_{\mathcal{R}}$ of 0.23 and 0.25 and a CMR of about 0.3 per image.}

\new{For all models, the error is similar at both granularities, whereas aggregating over 40~m segments, the granularity at which the risk event maps are generated, lowers the critical miss rate, since a segment is flagged as soon as one of its nodes reveals the hazard (from 0.29 to 0.21 for Qwen-VL and from 0.31 to 0.23 for Gemma-4). This removes only about a quarter of the critical misses, which come mainly from missed obstacles and surface hazards for Qwen-VL and from missed non-sidewalk paths for Gemma-4, suggesting that many of the remaining misses affect several consecutive nodes. Over-warnings, in turn, come mostly from false non-sidewalk alarms for Qwen-VL and false surface and non-sidewalk alarms for Gemma-4.}

\new{Processing cost is compatible with the offline setting: on a single RTX~4090 the mean time per image, including all triggered follow-up questions, goes from 0.2~s (ViLT) and 0.7~s (InstructBLIP) to 2.5~s for Gemma-4, 7.1~s for LLaVA and 23.7~s for Qwen-VL, that is, about 0.7~h versus 6.6~h to analyze 1,000 images with Gemma-4 and Qwen-VL, respectively. Since each map is computed once per location, this cost is paid only once, and Gemma-4 offers a favorable trade-off between risk-score quality and cost.}

\subsection{User Study}\label{sec:uc}

\new{To assess the perceived reliability of the proposed risk assessment framework under realistic conditions, we conducted a user evaluation using self-recorded Biel Glasses sequences from Barcelona (20 sequences, 20 nodes each). Evaluating complete sequences independently would result in an imbalanced assessment, as several routes contain predominantly safe scenes and rarely trigger Level-2 or Level-3 risk questions. To ensure a more balanced representation of risk conditions, we pooled the nodes from all sequences and organized them into three image sets, distributing Level-1, Level-2 and Level-3 risk cases as evenly as possible across the sets. Each set was presented through an online form to a different group of independent participants recruited from the general population, rather than specifically from people with low vision. For each image, participants were shown the image and the risk band predicted by Qwen-VL and asked whether they agreed with the assessment or would assign a different risk band. At the end of the form, participants rated how trustworthy they considered the model's risk assessments on a scale from 1 to 5. In total, 20 participants provided 300 assessments across 45 distinct images.}

\new{Participants agreed with the model's predicted risk band in 72.7\% of the assessments, while 96.7\% of their responses differed by at most one risk band. Agreement was highest for \textit{Safe} (87.2\%) and progressively lower for \textit{Low Risk} (73.2\%), \textit{Moderate} (66.7\%) and \textit{High Risk} (61.3\%), suggesting that distinguishing between neighboring risk bands is more challenging, particularly at the higher end of the risk spectrum. Disagreements tended to be conservative: participants assigned a lower risk band than the model in 54 of the 82 disagreements, while only one of the 78 images classified as \textit{Safe} by the model was judged to be \textit{Moderate}. Overall, participants rated the trustworthiness of the model's risk assessments at 4.1 out of 5, with 18 of the 20 participants assigning a rating of 4 or 5. These results provide an initial indication of agreement with the model's predictions and perceived trustworthiness. However, a comprehensive evaluation involving people with low vision, which is beyond the scope of this paper, will be necessary to further assess the practical relevance of the proposed risk assessments for the intended end users.}

\subsection{Limitations}\label{sec:lim}

\new{Our work has several limitations. First, although we evaluate the quality of the generated risk event maps, we do not assess how they influence route planning or navigation decisions. In particular, the potential benefits for route selection and the trade-off between safety and travel distance remain to be investigated. Second, hazard detection is still imperfect: even the best-performing models miss approximately 20\% of the annotated hazards. Consequently, the proposed maps should be regarded as supplementary decision-support tools rather than guarantees of safe passage. Third, risk event maps are generated offline from previously captured images and therefore reflect environmental conditions at the time of acquisition. Temporary hazards, such as roadworks or crowds, may consequently become outdated, as the current framework does not incorporate a mechanism for continuous map updates. We also have not systematically evaluated the effects of challenging environmental conditions, including nighttime, rain and visual occlusions. Furthermore, processing requirements increase with the geographical coverage of the mapped area, posing scalability challenges for city-wide deployment. In addition, the risk-score weights and severity thresholds were defined empirically and may benefit from further calibration using larger-scale feedback, particularly from people with low vision. Finally, risk perception is inherently subjective, meaning that the ground-truth annotations inevitably reflect the judgments of the annotators involved in this study.}
\section{Conclusions}\label{sec:conlcusions}

\noindent \new{We presented a framework for estimating pedestrian risk at the street-segment level to support safer navigation for people with low vision. The framework leverages VQA with a carefully designed three-level hierarchical query strategy to characterize urban hazards, combine their severity into normalized risk scores, and generate georeferenced risk event maps. We also introduced a geographically diverse dataset of 4,874 annotated images from 89 cities across all continents, comprising over 95,000 answered questions, and benchmarked five VQA models.}

\new{Results showed that generative MLLMs can detect pedestrian hazards without task-specific retraining, with Gemma-4 achieving the best overall detection performance (F1-score 0.71; balanced accuracy 0.82) and Qwen-VL showing particularly high recall for safety-critical hazards. However, fine-grained hazard characterization remains challenging, and risk estimation still produces critical misses, with approximately one in five dangerous segments classified as \textit{Safe} or \textit{Low Risk}. A preliminary user study showed 73\% agreement with the predicted risk bands and a mean trustworthiness rating of 4.1/5.}

\new{Future work will focus on improving hazard characterization and risk-score calibration through feedback from people with low vision, evaluating the impact of risk event maps on route planning, and enabling dynamic map updates as new observations become available.}

\section*{Acknowledgments}
\noindent Project CPP2021-008760 funded by MCIU/AEI /10.13039/501100011033 and by the ``European Union NextGenerationEU/PRTR''

\bibliographystyle{IEEEtran}
\bibliography{references}

\vspace{-0.5\baselineskip}
\begin{IEEEbiography}[{\includegraphics[width=1in,height=1.25in,clip,keepaspectratio]{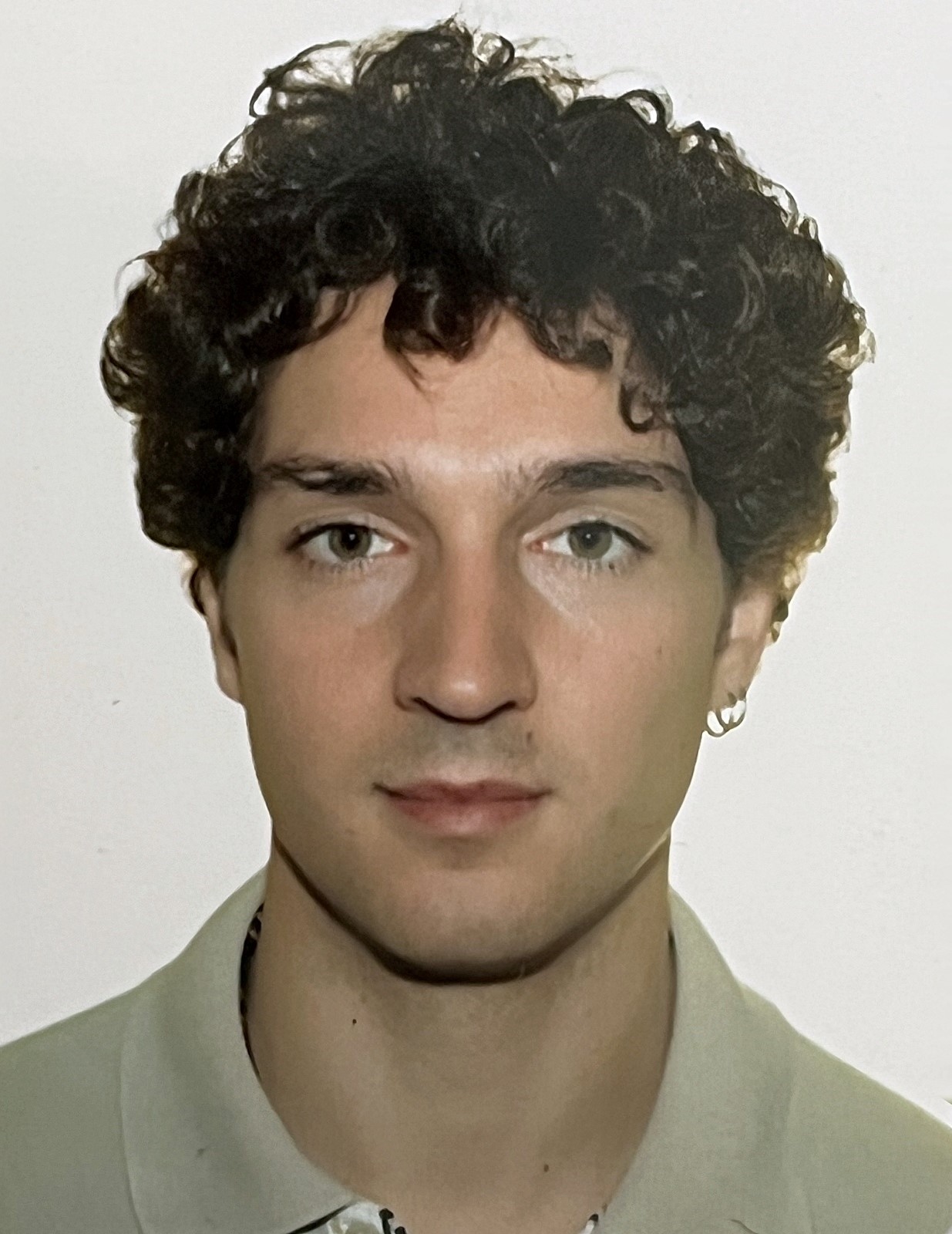}}]{Antoni Valls}
received the B.Sc. degree in theoretical physics from the 
Universitat de Barcelona, Spain, in 2022, and the 
M.Sc. degree in data science from the University of Padua,
Italy, in 2024. He has previously collaborated with the Institut d'Investigació en Intel·ligència Artificial (IIIA), CSIC, Barcelona, Spain, on natural language processing research, and worked as a Robotic Engineer with the Institut de Robòtica i Informàtica Industrial (IRI), CSIC-UPC, Barcelona, Spain, on the SMARTGAZE~II project in collaboration with Biel Glasses, developing AI-powered assistive systems to enhance the mobility of people with low vision. He is currently a Computer Vision Engineer with Biel Glasses, Barcelona, Spain, while pursuing the Ph.D. degree at the Universitat Politècnica de Catalunya (UPC), Barcelona, Spain. His research interests include computer vision, vision-language models, assistive and pedestrian navigation, and localization in dynamic environments.
\end{IEEEbiography}
\begin{IEEEbiography}[{\includegraphics[width=1in,height=1.25in,clip,keepaspectratio]{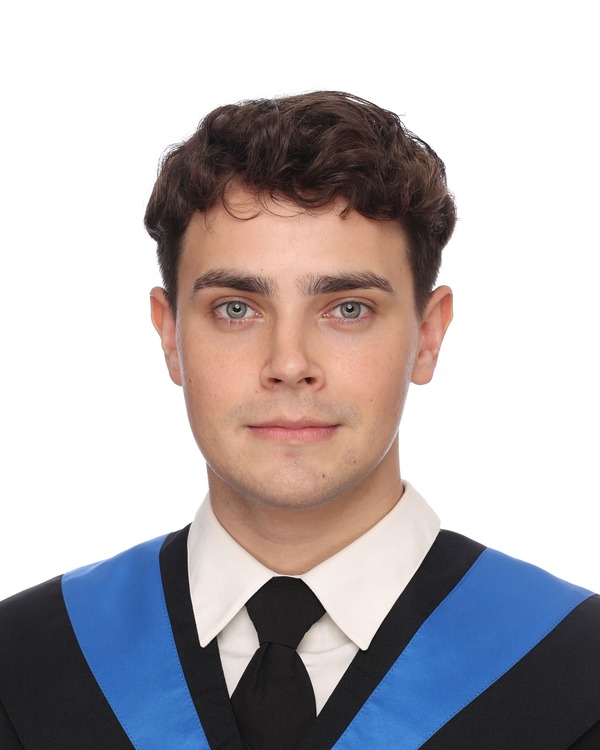}}]{Jorge Tyrakowski}
received the B.S. degree in computer science from National Yang Ming Chiao Tung University (NYCU), Hsinchu, Taiwan, in 2026. He is currently working toward the M.S. degree in computer science and information engineering at National Taiwan University (NTU), Taipei, Taiwan, where he is a member of the AIMM Research Group, Communications and Multimedia Laboratory. Previously, he was a research assistant with the Applied Computing and Multimedia Laboratory, NYCU, and an R\&D intern with the Industrial Technology Research Institute (ITRI), Hsinchu, Taiwan, where he worked on deploying LLMs and VLMs on edge devices. His research interests include vision-language models, efficient inference of multimodal large language models, and edge AI.
\end{IEEEbiography}
\begin{IEEEbiography}[{\includegraphics[width=1in,height=1.25in,clip,keepaspectratio]{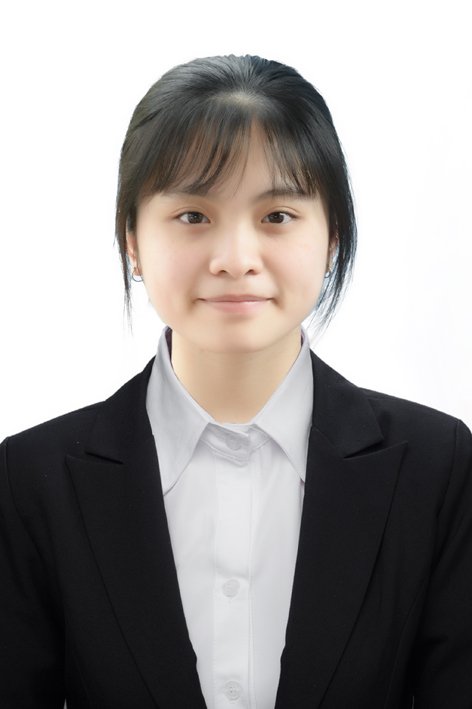}}]{My-Dinh Le}
received the B.S. degree in Computer Science from National Dong Hwa University (NDHU), Taiwan, in 2026. She is currently enrolled in the M.S. program in Data Science at National Taiwan University (NTU), Taiwan. She is a member of the AIMM Research Group, Communications and Multimedia Laboratory. During university, she was a research assistant for the Machine Perception and Learning Laboratory at NDHU. Her research interests include Computer Vision application in healthcare, Human Activity Recognition, and Adversarial attack.
\end{IEEEbiography}
\begin{IEEEbiography}[{\includegraphics[width=1in,height=1.25in,clip,keepaspectratio]{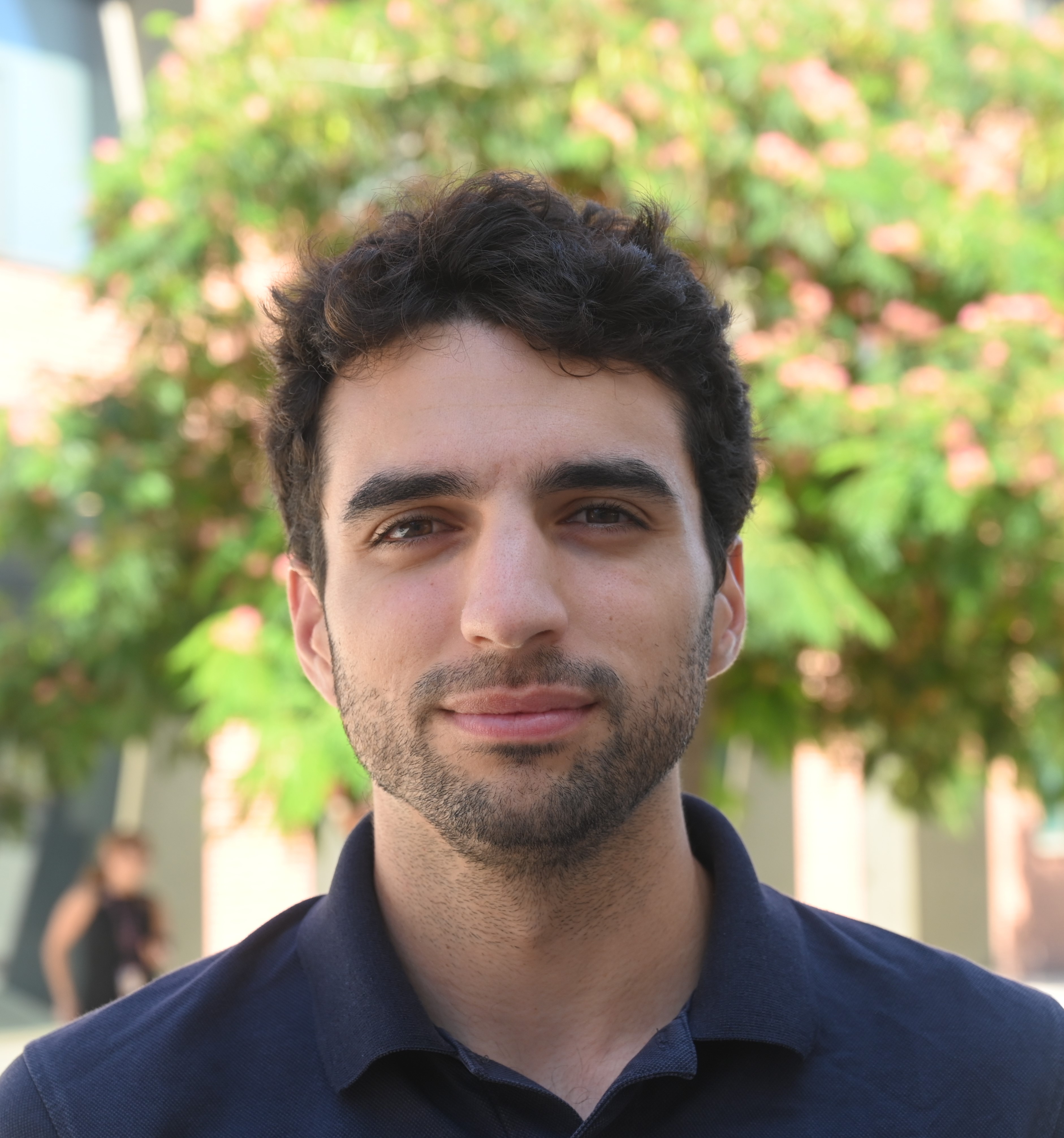}}]{Alex Cabañeros}
received the B.S. degree in industrial electronics and automatic control engineering in 2014 and the M.Sc. degree in automatic control and robotics in 2017, both from the Universitat Politècnica de Catalunya (UPC), Spain. During his master’s studies, he specialized in computer vision and machine learning. In 2017, he collaborated with Eindhoven University of Technology (TU/e), the Netherlands, on the ROPOD project. He is currently the Chief Operating Officer of Biel Glasses S.L., Barcelona, Spain, where he oversees the development of a wearable assistive device designed to support the mobility of people with low vision. He contributed to SMARTGAZE II (CPP2021-008760), a collaboration between Biel Glasses S.L., IRI (CSIC-UPC), and LEITAT under Spain’s Proyectos de Colaboración Público-Privada program. The research presented in this paper was conducted as part of SMARTGAZE II. His research interests include computer vision, assistive navigation, and the design and validation of wearable technologies for people with low vision.
\end{IEEEbiography}
\begin{IEEEbiography}[{\includegraphics[width=1in,height=1.25in,clip,keepaspectratio]{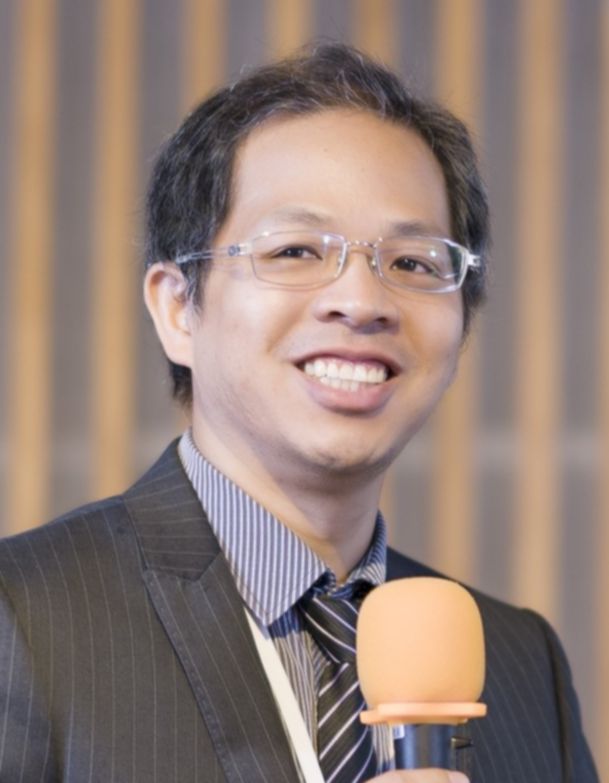}}]{Wen-Huang Cheng} is a University Distinguished Chair Professor and Chair of the Department of Computer Science and Information Engineering at National Taiwan University, and an Affiliate Faculty at VinUniversity. His research interests include artificial intelligence, multimedia, computer vision, and machine learning. He has held leadership roles in major journals and conferences, including Senior Editor of IEEE Consumer Electronics Magazine and IEEE Transactions on Consumer Electronics, Associate Editor of IEEE TPAMI and TMM, and General or Technical Program Chair of ACM Multimedia, IEEE ICME, ACM ICMR, and ACM MMAsia. He has also served as a Governing Board Member of IAPR. His honors include the NVIDIA Academic Grant Program Award, the NSTC Outstanding Research Award, the IEEE Consumer Electronics Magazine Best Paper Award, the IEEE ICME Best Paper Award, and the IEEE TMM Outstanding Associate Editor Award. He is an IEEE Fellow, IAPR Fellow, IET Fellow, and ACM Distinguished Member.
\end{IEEEbiography}
\begin{IEEEbiography}[{\includegraphics[width=1in,height=1.25in,clip,keepaspectratio]{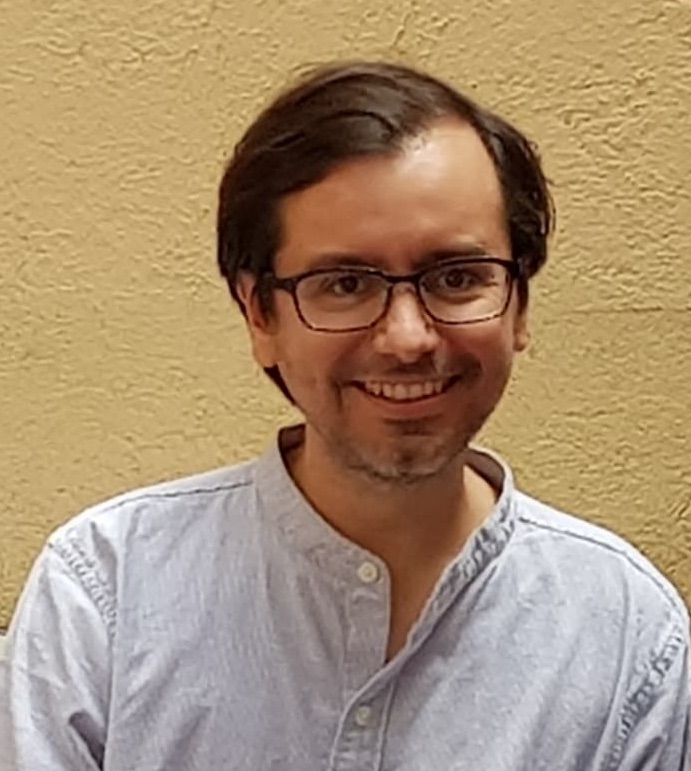}}]
{Jordi Sanchez-Riera} is an associate researcher at the Spanish Scientific Research Council (CSIC) in Barcelona. He earned his B.S. in computer science and M.S. in electronic engineering from UPC, and a Ph.D. from the University of Grenoble. He has over 15 years of research experience in computer vision, machine learning, AI and robotics, with interests in 3D vision, human reconstruction, multimodal perception, generative AI and robotic perception. He has published extensively in leading international venues and conducted research stays in the US, Switzerland and Taiwan. His distinctions include a Google Scholarship, an Academia Sinica Postdoctoral Fellowship and a Juan de la Cierva–Incorporación grant.
\end{IEEEbiography}

\end{document}